%% file: 79.tex
\documentclass{llncs}
\usepackage[utf8]{inputenc}
\usepackage[nolist]{acronym}
\usepackage{amsfonts,epsfig,latexsym,amssymb,yhmath,hyperref}
\usepackage{booktabs}
\usepackage{array}
\usepackage{makeidx}  
\usepackage[usenames,dvipsnames]{color}

\newcolumntype{L}[1]{>{\raggedright\let\newline\\\arraybackslash\hspace{0pt}}m{#1}}
\newcolumntype{C}[1]{>{\centering\let\newline\\\arraybackslash\hspace{0pt}}m{#1}}
\newcolumntype{R}[1]{>{\raggedleft\let\newline\\\arraybackslash\hspace{0pt}}m{#1}}

\usepackage{fancyhdr}
\fancypagestyle{firststyle}
{
   \fancyhf{}
   \lhead{To be published in the proceedings of IPMI 2017}
}

\title{
Unsupervised Anomaly Detection with Generative Adversarial Networks to Guide Marker Discovery
}

\author{Thomas Schlegl\inst{1,2}
\thanks{This work has received funding from IBM, FWF (I2714-B31), OeNB (15356, 15929), the Austrian Federal Ministry of Science, Research and Economy (CDL OPTIMA).},
Philipp Seeb\"ock\inst{1,2} \and Sebastian M. Waldstein\inst{2} \and Ursula~Schmidt-Erfurth\inst{2} \and Georg Langs\inst{1}
}
\institute{$^1$Computational Imaging Research Lab, Department of Biomedical Imaging and Image-guided Therapy, Medical University Vienna, Austria
\email{thomas.schlegl@meduniwien.ac.at}\\
$^2$Christian Doppler Laboratory for Ophthalmic Image Analysis, Department of Ophthalmology and Optometry, Medical University Vienna, Austria}

\begin{document}

\maketitle
\thispagestyle{firststyle}

\begin{abstract}
Obtaining models that capture imaging markers relevant for disease progression and treatment monitoring is challenging. Models are typically based on large amounts of data with annotated examples of known markers aiming at automating  detection. High annotation effort and the limitation to a vocabulary of known markers limit the power of such approaches. Here, we perform unsupervised learning to identify anomalies in imaging data as candidates for markers. We propose \textit{AnoGAN}, a deep convolutional generative adversarial network to learn a manifold of normal anatomical variability, accompanying a novel anomaly scoring scheme based on the mapping from image space to a latent space. Applied to new data, the model labels anomalies, and scores image patches indicating their fit into the learned distribution. Results on optical coherence tomography images of the retina demonstrate that the approach correctly identifies anomalous images, such as images containing retinal fluid or hyperreflective foci.

\end{abstract}

\input{introduction.tex}
\input{method.tex}

\input{experiments.tex}

\input{discussion.tex}

\bibliographystyle{splncs}
\bibliography{references}
\end{document}

%% file: introduction.tex
\section{Introduction}

The detection and quantification of disease markers in imaging data is critical during diagnosis, and monitoring of disease progression, or treatment response. Relying on the vocabulary of known markers limits the use of imaging data containing far richer relevant information. Here, we demonstrate that relevant \emph{anomalies} can be identified by unsupervised learning on large-scale imaging data.

Medical imaging enables the observation of markers correlating with disease status, and treatment response. While there is a wide range of known markers (e.g., characteristic image appearance of brain tumors or calcification patterns in breast screening), many diseases lack a sufficiently broad set,
while in others the predictive power of markers is limited.  Furthermore, even if predictive markers are known, their computational detection in imaging data typically requires extensive supervised training using large amounts of annotated data such as labeled lesions. This limits our ability to exploit imaging data for treatment decisions. 

Here, we propose unsupervised learning to create a rich generative model of healthy local anatomical appearance. We show how generative adversarial networks (GANs) can solve the central problem of creating a sufficiently representative model of appearance, while at the same time learning a generative and discriminative component. We propose an improved technique for mapping from image space to latent space. We use both components to differentiate between observations that conform to the training data and such data that does not fit.

\begin{figure}[tp]
 \centering
     \includegraphics[width=1\textwidth]{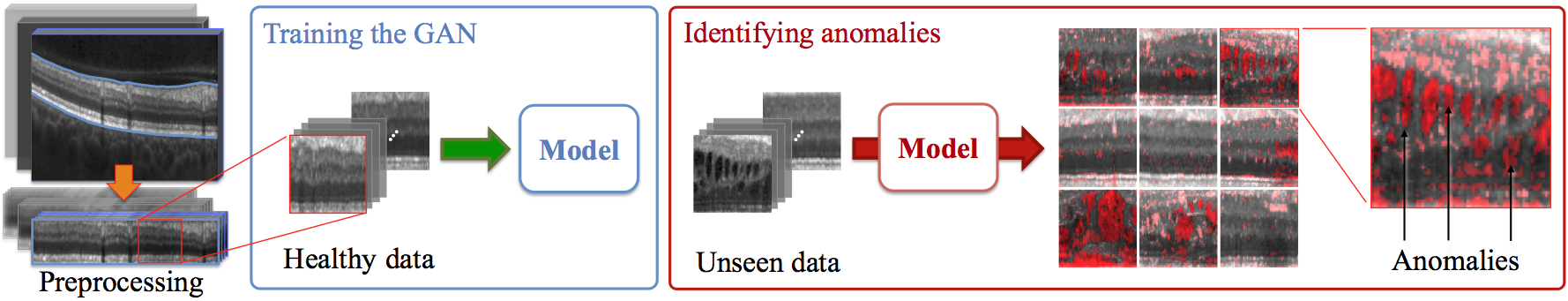}
    \vspace{-3mm}
  \caption{Anomaly detection framework. The preprocessing step includes extraction and flattening of the retinal area, patch extraction and intensity normalization. Generative adversarial training is performed on healthy data and testing is performed on both, unseen healthy cases and anomalous data.}
\label{fig:method/overview}

\vspace{-5mm}
\end{figure}

\paragraph{Related Work}

Anomaly detection is the task of identifying test data not fitting the \emph{normal} data distribution seen during training. Approaches for anomaly detection exist in various domains, ranging from video analysis~\cite{del2016discriminative} to remote sensing~\cite{matteoli2014overview}. They typically either use an explicit representation of the distribution of normal data in a feature space, and determine outliers based on the local density at the observations' position in the feature space. Carrera et al.~\cite{carrera2015detecting} utilized convolutional sparse models to learn a dictionary of filters to detect anomalous regions in texture images. Erfani et al.~\cite{erfani2016high} proposed a hybrid model for unsupervised anomaly detection that uses a one-class support vector machine (SVM). The SVM was trained from features that were learned by a deep belief network (DBN). The experiments in the aforementioned works were performed on real-life-datasets comprising 1D inputs, synthetic data or texture images, which have lower dimensionality or different data characteristics compared to medical images. An investigation of anomaly detection research papers can be found in~\cite{pimentel2014review}. In clinical optical coherence tomography (OCT) scan analysis, Venhuizen et al.~\cite{venhuizen2015automated} used bag-of-word features as a basis for supervised random forest classifier training to distinguish diseased patients from healthy subjects. Schlegl et al.~\cite{schlegl2015ipmi} utilized convolutional neural networks to segment retinal fluid regions in OCT data via weakly supervised learning based on semantic descriptions of pathology-location pairs extracted from medical reports. In contrast to our approach, both works used some form of supervision for classifier training. Seeb\"ock et al.~\cite{seebock2016identifying} identified anomalous regions in OCT images through unsupervised learning on healthy examples, using a convolutional autoencoder and a one-class SVM, and explored different classes of anomalies. In contrast to this work, the SVM in~\cite{seebock2016identifying} involved the need to choose a hyper-parameter that defined the amount of training points covered by the estimated healthy region.

GANs enable to learn generative models generating detailed realistic images~\cite{goodfellow2014generative,denton2015deep,donahue2016adversarial}. Radford et al.~\cite{radford2015unsupervised} introduced deep convolutional generative adversarial networks (DCGANs) and showed that GANs are capable of capturing semantic image content enabling vector arithmetic for visual concepts.
Yeh et al.~\cite{yeh2016semantic} trained GANs on natural images and applied the trained model for semantic image inpainting. Compared to Yeh et al.~\cite{yeh2016semantic}, we implement two adaptations for an improved mapping from images to the latent space.
We condition the search in the latent space on the whole query
image, and propose a novel variant to guide the search in the latent space (inspired by feature matching~\cite{salimans2016improved}). In addition, we define an anomaly score, which is not needed in an inpainting task. The main difference of this paper to aforementioned anomaly detection work is the representative power of the generative model and the coupled mapping schema, which utilizes a trained DCGAN and enables accurate discrimination between normal anatomy, and local anomalous appearance. This renders the detection of subtle anomalies at scale feasible.

\paragraph{Contribution}

In this paper, we propose adversarial training of a generative model of normal appearance (see blue block in Figure~\ref{fig:method/overview}), described in Section~\ref{sec:methods:gan_training}, and a coupled mapping schema, described in Section~\ref{sec:methods:mapping}, that enables the evaluation of novel data (Section ~\ref{sec:methods:ano_detect}) to identify anomalous images and segment anomalous regions within imaging data (see red block in Figure~\ref{fig:method/overview}).
Experiments on labeled test data, extracted from spectral-domain OCT (SD-OCT) scans, show that this approach identifies known anomalies with high accuracy, and at the same time detects other anomalies for which no voxel-level annotations are available.
To the best of our knowledge, this is the first work, where GANs are used for anomaly or novelty detection. Additionally, we propose a novel mapping approach, wherewith the pre-image problem can be tackled.

%% file: method.tex
\section{Generative Adversarial Representation Learning to Identify Anomalies}

\begin{figure}[tp]
 \centering
    \includegraphics[width=1\textwidth]{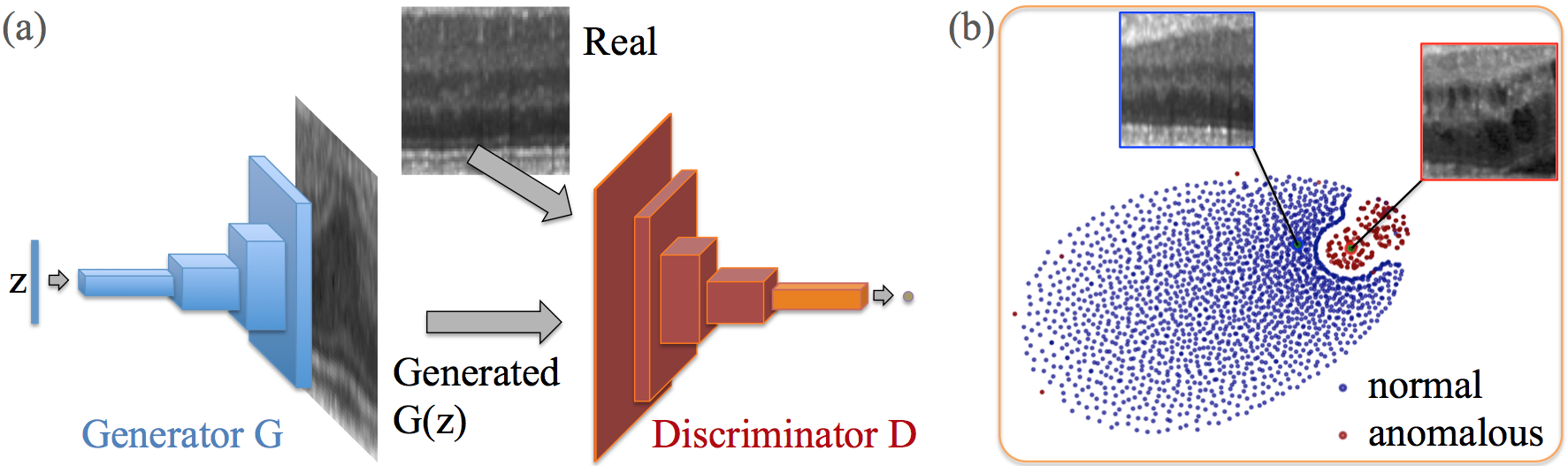}
  \caption{(a) Deep convolutional generative adversarial network. (b) t-SNE embedding of normal (blue) and anomalous (red) images on the feature representation of the last convolution layer (orange in (a)) of the discriminator.}
\label{fig:method/gan}

\vspace{-5mm}
\end{figure}

To identify anomalies, we learn a model representing normal anatomical variability based on GANs~\cite{yeh2016semantic}. This method trains a generative model, and a discriminator to distinguish between generated and real data simultaneously (see Figure~\ref{fig:method/gan}(a)). Instead of a single cost function optimization, it aims at the Nash equilibrium of costs, increasing the representative power and specificity of the generative model, while at the same time becoming more accurate in classifying real- from generated data and improving the corresponding feature mapping. In the following we explain how to build this model (Section~\ref{sec:methods:gan_training}), and how to use it to identify appearance not present in the training data (Sections~\ref{sec:methods:mapping} and~\ref{sec:methods:ano_detect}).

\subsection{Unsupervised Manifold Learning of Normal Anatomical Variability}
\label{sec:methods:gan_training}
We are given a set of $M$ medical images $\mathbf{I}_m$ showing healthy anatomy, with $m=1,2,\dots ,M$, where $\mathbf{I}_m \in \mathbb{R}^{a \times b}$ is an intensity image of size $a \times b$. From each image $\mathbf{I}_m$, we extract $K$ 2D image patches $x_{k,m}$ of size $c \times c$ from randomly sampled positions resulting in data $\mathbf{x}=x_{k,m} \in \mathcal{X}$, with $k=1,2,\dots ,K$. During training we are only given $\langle \mathbf{I}_m \rangle$ and train a generative adversarial model to learn the manifold $\mathcal{X}$ (blue region in Figure~\ref{fig:method/gan}(b)), which represents the variability of the training images, in an unsupervised fashion.
For testing, we are given $\langle \mathbf{y}_n, l_n\rangle$, where $\mathbf{y}_n$ are unseen images of size $c \times c$ extracted from new testing data $\mathbf{J}$ and $l_n \in \{0,1\}$ is an array of binary image-wise ground-truth labels, with $n=1,2,\dots ,N$. These labels are only given during testing, to evaluate the anomaly detection performance based on a given pathology.

\paragraph{Encoding Anatomical Variability with a Generative Adversarial Network.}
A GAN consists of two adversarial modules, a generator $G$ and a discriminator $D$. The generator $G$ learns a distribution $p_g$ over data $\mathbf{x}$ via a mapping $G(\mathbf{z})$ of samples $\mathbf{z}$, 1D vectors of uniformly distributed input noise sampled from latent space $\mathcal{Z}$, to 2D images in the image space manifold $\mathcal{X}$, which is populated by healthy examples. In this setting, the network architecture of the generator $G$ is equivalent to a convolutional decoder that utilizes a stack of strided convolutions.
The discriminator $D$ is a standard CNN that maps a 2D image to a single scalar value $D(\cdot)$. The discriminator output $D(\cdot)$ can be interpreted as probability that the given input to the discriminator $D$ was a real image $\mathbf{x}$ sampled from training data $\mathcal{X}$ or generated $G(\mathbf{z})$ by the generator $G$.
$D$ and $G$ are simultaneously optimized through the following two-player minimax game with value function $V(G, D)$~\cite{goodfellow2014generative}:
\begin{equation}\label{eqn:loss}
	\underset{G}{\operatorname{min}}\, \underset{D}{\operatorname{max}}\, V(D,G) = \mathbb{E}_{\mathbf{x} \sim p_{data}(\mathbf{x})} \left[\log D(\mathbf{x}) \right] + \mathbb{E}_{\mathbf{z} \sim p_{\mathbf{z}}(\mathbf{z})} \left[\log (1 - D(G(\mathbf{z}))) \right].
\end{equation}
The discriminator is trained to maximize the probability of assigning real training examples the \textit{``real''} and samples from $p_g$ the \textit{``fake''} label.
The generator $G$ is simultaneously trained to fool $D$ via minimizing $V(G) = \log(1 - D(G(\mathbf{z})))$,
which is equivalent to maximizing
\begin{equation}\label{eqn:LossGmax}
	V(G) = D(G(\mathbf{z})).
\end{equation}

During adversarial training the generator improves in generating realistic images and the discriminator progresses in correctly identifying real and generated images.

\subsection{Mapping new Images to the Latent Space} \label{sec:methods:mapping}
When adversarial training is completed, the generator has learned the mapping $G(\mathbf{z}) = \mathbf{z} \mapsto \mathbf{x}$ from latent space representations $\mathbf{z}$ to realistic (normal) images $\mathbf{x}$.
But GANs do not automatically yield the inverse mapping $\mu(\mathbf{x}) = \mathbf{x} \mapsto \mathbf{z}$ for free. The latent space has smooth transitions~\cite{radford2015unsupervised}, so sampling from two points close in the latent space generates two visually similar images. Given a query image $\mathbf{x}$, we aim to find a point $\mathbf{z}$ in the latent space that corresponds to an image $G(\mathbf{z})$ that is visually most similar to query image $\mathbf{x}$ and that is located on the manifold $\mathcal{X}$.
The degree of similarity of $\mathbf{x}$ and $G(\mathbf{z})$ depends on to which extent the query image follows the data distribution $p_g$ that was used for training of the generator.
To find the best $\mathbf{z}$, we start with randomly sampling $\mathbf{z}_1$ from the latent space distribution $\mathcal{Z}$ and feed it into the trained generator to get a generated image $G(\mathbf{z}_1)$. Based on the generated image $G(\mathbf{z}_1)$ we define a loss function, which provides gradients for the update of the coefficients of $\mathbf{z}_1$ resulting in an updated position in the latent space, $\mathbf{z}_2$. In order to find the most similar image $G(\mathbf{z}_{\Gamma})$, the location of $\mathbf{z}$ in the latent space $\mathcal{Z}$ is optimized in an iterative process via $\gamma=1,2,\dots ,\Gamma$ backpropagation steps.

In the spirit of~\cite{yeh2016semantic}, we define a loss function for the mapping of new images to the latent space that comprises two components, a \textit{residual loss} and a \textit{discrimination loss}. The \textit{residual loss} enforces the visual similarity between the generated image $G(\mathbf{z}_{\gamma})$ and query image $\mathbf{x}$. The \textit{discrimination loss} enforces the generated image $G(\mathbf{z}_{\gamma})$ to lie on the learned manifold $\mathcal{X}$.
Therefore, both components of the trained GAN, the discriminator $D$ and the generator $G$, are utilized to adapt the coefficients of $\mathbf{z}$ via backpropagation. In the following, we give a detailed description of both components of the loss function.

\subsubsection{Residual Loss}
The \textit{residual loss} measures the visual dissimilarity between query image $\mathbf{x}$ and generated image $G(\mathbf{z}_{\gamma})$ in the image space and is defined by
\begin{equation}\label{eqn:resid_loss}
	\mathcal{L}_{R}(\mathbf{z}_{\gamma}) = \sum | \mathbf{x} - G(\mathbf{z}_{\gamma}) |.
\end{equation}

Under the assumption of a perfect generator $G$ and a perfect mapping to latent space, for an ideal normal query case, images $\mathbf{x}$ and $G(\mathbf{z}_{\gamma})$ are identical. In this case, the \textit{residual loss} is zero.

\subsubsection{Discrimination Loss}
For image inpainting, Yeh et al.~\cite{yeh2016semantic} based the computation of the \textit{discrimination loss} $\mathcal{L}_{\hat{D}}(\mathbf{z}_{\gamma})$ on the discriminator output by feeding the generated image $G(\mathbf{z}_{\gamma})$ into the discriminator $\mathcal{L}_{\hat{D}}(\mathbf{z}_{\gamma}) = \sigma(D(G(\mathbf{z}_{\gamma})), \alpha)$, where $\sigma$ is the sigmoid cross entropy, which defined the discriminator loss of real images during adversarial training, with logits $D(G(\mathbf{z}_{\gamma}))$ and targets $\alpha=1$.

\subsubsection{An improved discrimination loss based on feature matching}
In contrast to the work of Yeh et al.~\cite{yeh2016semantic}, where $\mathbf{z}_{\gamma}$ is updated to fool $D$, we define an alternative discrimination loss $\mathcal{L}_{D}(\mathbf{z}_{\gamma})$, where $\mathbf{z}_{\gamma}$ is updated to match $G(\mathbf{z}_{\gamma})$ with the learned distribution of normal images. This is inspired by the recently proposed feature matching technique~\cite{salimans2016improved}.

Feature matching addresses the instability of GANs due to over-training on the discriminator response~\cite{salimans2016improved}. In the feature matching technique, the objective function for optimizing the generator is adapted to improve GAN training. Instead of optimizing the parameters of the generator via maximizing the discriminator's output on generated examples (Eq.~\eqref{eqn:LossGmax}), the generator is forced to generate data that has similar statistics as the training data, i.e. whose intermediate feature representation is similar to those of real images.
Salimans et al.~\cite{salimans2016improved} found that feature matching is especially helpful when classification is the target task. Since we do not use any labeled data during adversarial training, we do not aim for learning class-specific discriminative features but we aim for learning good representations. Thus, we do not adapt the training objective of the generator during adversarial training, but instead use the idea of feature matching to improve the mapping to the latent space. Instead of using the scalar output of the discriminator for computing the \textit{discrimination loss}, we propose to use a richer intermediate feature representation of the discriminator and define the \textit{discrimination loss} as follows:
\begin{equation}\label{eqn:discr_loss_feat}
	\mathcal{L}_{D}(\mathbf{z}_{\gamma}) = \sum | \mathbf{f(\mathbf{x})} - \mathbf{f}(G(\mathbf{z}_{\gamma})) |,
\end{equation}
where the output of an intermediate layer $f(\mathbf{\cdot})$ of the discriminator is used to specify the statistics of an input image. Based on this new loss term, the adaptation of the coordinates of $\mathbf{z}$ does not only rely on a hard decision of the trained discriminator, whether or not a generated image $G(\mathbf{z}_{\gamma})$ fits the learned distribution of normal images, but instead takes the rich information of the feature representation, which is learned by the discriminator during adversarial training, into account. In this sense, our approach utilizes the trained discriminator not as classifier but as a feature extractor.\newline

For the mapping to the latent space, we define the overall loss as weighted sum of both components:
\begin{equation}\label{eqn:overall_loss}
	\mathcal{L}(\mathbf{z}_{\gamma}) = (1-\lambda) \cdot \mathcal{L}_{R}(\mathbf{z}_{\gamma}) + \lambda \cdot \mathcal{L}_{D}(\mathbf{z}_{\gamma}).
\end{equation}
Only the coefficients of $\mathbf{z}$ are adapted via backpropagation. The trained parameters of the generator and discriminator are kept fixed.

\subsection{Detection of Anomalies}
\label{sec:methods:ano_detect}
During anomaly identification in new data we evaluate the new query image $\mathbf{x}$ as being a normal or anomalous image. Our loss function (Eq.~\eqref{eqn:overall_loss}), used for mapping to the latent space, evaluates in every update iteration $\gamma$ the compatibility of generated images $G(\mathbf{z}_{\gamma})$ with images, seen during adversarial training. Thus, an \textit{anomaly score}, which expresses the fit of a query image $\mathbf{x}$ to the model of normal images, can be directly derived from the mapping loss function (Eq.~\eqref{eqn:overall_loss}):
\begin{equation}\label{eqn:overall_score}
	A(\mathbf{x}) =  (1-\lambda) \cdot R(\mathbf{x}) + \lambda \cdot D(\mathbf{x}),
\end{equation}
where the \textit{residual score} $R(x)$ and the \textit{discrimination score} $D(x)$ are defined by the \textit{residual loss} $\mathcal{L}_{R}(\mathbf{z}_{\Gamma})$ and the \textit{discrimination loss} $\mathcal{L}_{D}(\mathbf{z}_{\Gamma})$ at the last ($\Gamma^{th}$) update iteration of the mapping procedure to the latent space, respectively. The model yields a large \textit{anomaly score} $A(\mathbf{x})$ for anomalous images, whereas a small \textit{anomaly score} means that a very similar image was already seen during training. We use the \textit{anomaly score} $A(\mathbf{x})$ for image based anomaly detection. Additionally, the residual image $\mathbf{x}_R = | \mathbf{x} - G(\mathbf{z}_{\Gamma}) |$ is used for the identification of anomalous regions within an image. For purposes of comparison, we additionally define a \textit{reference anomaly score} $\hat{A}(\mathbf{x}) =  (1-\lambda) \cdot R(\mathbf{x}) + \lambda \cdot \hat{D}(\mathbf{x})$, where $\hat{D}(\mathbf{x}) = \mathcal{L}_{\hat{D}}(\mathbf{z}_{\Gamma})$ is the 
\textit{reference discrimination score} used by Yeh et al.~\cite{yeh2016semantic}.

%% file: experiments.tex
\section{Experiments}

\paragraph{Data, Data Selection and Preprocessing}
We evaluated the method on clinical high resolution SD-OCT volumes of the retina with 49 B-scans (representing an image slice in zx-plane) per volume and total volume resolutions of $496 \times 512 \times 49$ voxels in z-, x-, and y direction, respectively. The GAN was trained on 2D image patches extracted from 270 clinical OCT volumes of healthy subjects, which were chosen based on the criterion that the OCT volumes do not contain fluid regions. For testing, patches were extracted from 10 additional healthy cases and 10 pathological cases, which contained retinal fluid.
The OCT volumes were preprocessed in the following way. The gray values were normalized to range from -1 to 1. The volumes were resized in x-direction to a size of $22\mu m$ resulting in approximately 256 columns. The retinal area was extracted and flattened to adjust for variations in orientation, shape and thickness. We used an automatic layer segmentation algorithm following \cite{garvin2009automated} to find the top and bottom layer of the retina that define the border of the retina in z-direction. From these normalized and flattened volumes, we extracted in total 1.000.000 2D training patches with an image resolution of $64 \times 64$ pixels at randomly sampled positions. Raw data and preprocessed image representation are shown in Figure~\ref{fig:method/overview}.
The test set in total consisted of 8192 image patches and comprised normal and pathological samples from cases not included in the training set. For pathological OCT scans, voxel-wise annotations of fluid and non-fluid regions from clinical retina experts were available. These annotations were only used for statistical evaluation but were never fed to the network, neither during training nor in the evaluation phase. For the evaluation of the detection performance, we assigned a positive label to an image, if it contained at least a single pixel annotated as retinal fluid.

\paragraph{Evaluation}
The manifold of normal images was solely learned on image data of healthy cases with the aim to model the variety of healthy appearance. For performance evaluation in anomaly detection we ran the following experiments.

\noindent
\textbf{(1)} We explored qualitatively whether the model can generate realistic images. This assessment was performed on image patches of healthy cases extracted from the training set or test set and on images of diseased cases extracted from the test set.

\noindent
\textbf{(2)} We evaluated quantitatively the anomaly detection accuracy of our approach on images extracted from the annotated test set.
We based the anomaly detection on the \textit{anomaly score} $A(\mathbf{x})$ or only on one of both components, on the \textit{residual score} $R(\mathbf{x})$ or on the \textit{discrimination score} $D(\mathbf{x})$ and report receiver operating characteristic (ROC) curves of the corresponding anomaly detection performance on image level.

Based on our proposed \textit{anomaly score} $A(\mathbf{x})$, we evaluated qualitatively the segmentation performance and if additional anomalies were identified.

\noindent
\textbf{(3)} To provide more details of individual components' roles, and the gain by the proposed approach, we evaluated the effect on the anomaly detection performance, when for manifold learning the adversarial training is not performed with a DCGAN but with an adversarial convolutional autoencoder (aCAE)~\cite{DBLP:journals/corr/PathakKDDE16}, while leaving the definition of the \textit{anomaly score} unchanged. An \textit{aCAE} also implements a discriminator but replaces the generator by an encoder-decoder pipeline. The depth of the components of the trained \textit{aCAE} was comparable to the depth of our adversarial model.
As a second alternative approach, denoted as $GAN_R$, we evaluated the anomaly detection performance, when the \textit{reference anomaly score} $\hat{A}(\mathbf{x})$, or the \textit{reference discrimination score} $\hat{D}(\mathbf{x})$ were utilized for anomaly scoring and the corresponding losses were used for the mapping from image space to latent space, while the pre-trained GAN parameters of the \textit{AnoGAN} were used. We report ROC curves for both alternative approaches. Furthermore, we calculated sensitivity, specificity, precision, and recall at the optimal cut-off point on the ROC curves, identified through the Youden's index and report results for the \textit{AnoGan} and for both alternative approaches.

\paragraph{Implementation details}
As opposed to historical attempts, Radford et al.~\cite{radford2015unsupervised} identified a DCGAN architecture that resulted in stable GAN training on images of sizes $64 \times 64$ pixels. Hence, we ran our experiments on image patches of the same size and used widley the same DCGAN architecture for GAN training (Section~\ref{sec:methods:gan_training}) as proposed by Radford et al. ~\cite{radford2015unsupervised}\footnote{We adapted: https://github.com/bamos/dcgan-completion.tensorflow}. We used four fractionally-strided convolution layers in the generator, and four convolution layers in the discriminator, all filters of sizes $5 \times 5$. Since we processed gray-scale images, we utilized intermediate representations with $512-256-128-64$ channels (instead of $1024-512-256-128$ used in~\cite{radford2015unsupervised}). DCGAN training was performed for 20 epochs utilizing Adam~\cite{kingma2014adam}, a stochastic optimizer. We ran 500 backpropagation steps for the mapping (Section~\ref{sec:methods:mapping}) of new images to the latent space. We used $\lambda=0.1$ in Equations ~\eqref{eqn:overall_loss} and ~\eqref{eqn:overall_score}, which was found empirically due to preceding experiments on a face detection dataset. All experiments were performed using Python 2.7 with the TensorFlow~\cite{tensorflow2015-whitepaper} library and run on a Titan X graphics processing unit using CUDA 8.0. 

\subsection{Results}
Results demonstrate the generative capability of the DCGAN and the appropriateness of our proposed mapping and scoring approach for anomaly detection. We report qualitative and quantitative results on segmentation performance and detection performance of our approach, respectively.

\begin{figure}[tp]
  \centering
     \includegraphics[width=1\textwidth]{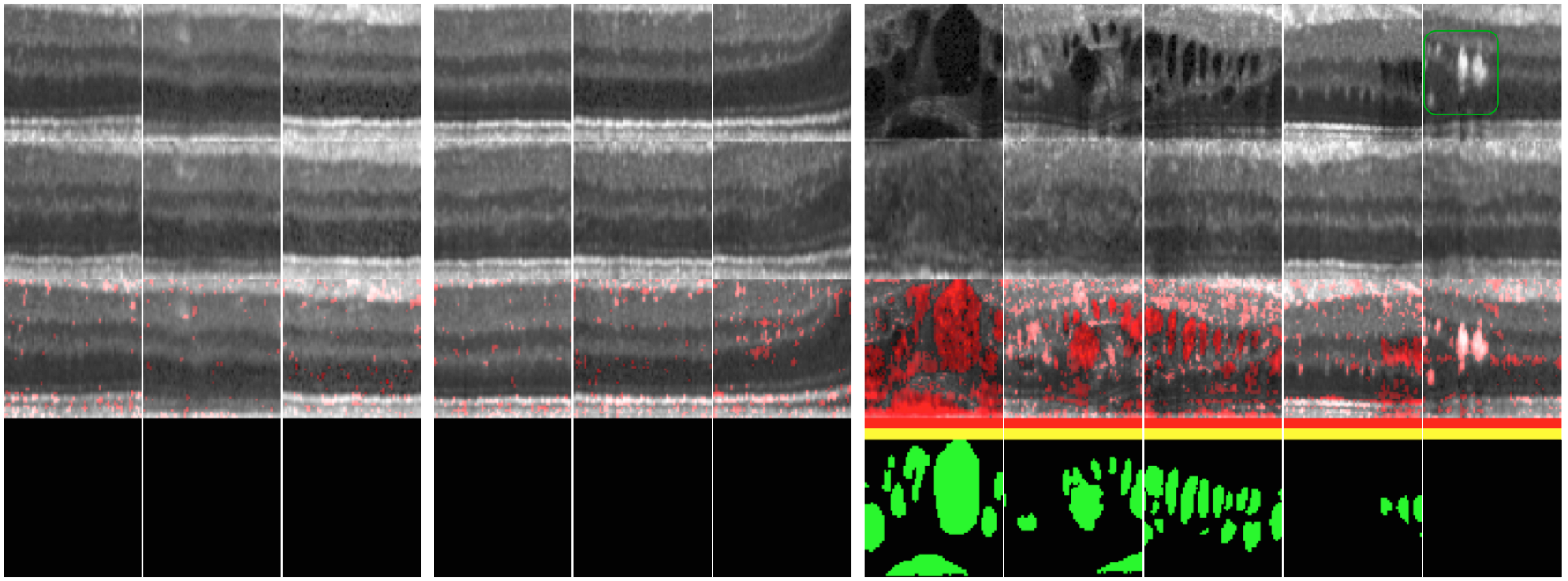}
  \caption{Pixel-level identification of anomalies on exemplary images. First row: Real input images. Second row: Corresponding images generated by the model triggered by our proposed mapping approach. Third row: Residual overlay. Red bar: Anomaly identification by \textit{residual score}. Yellow bar: Anomaly identification by \textit{discrimination score}. Bottom row: Pixel-level annotations of retinal fluid. First block and second block: Normal images extracted from OCT volumes of healthy cases in the training set and test set, respectively. Third block: Images extracted from diseased cases in the test set. Last column: Hyperreflective foci (within green box). (Best viewed in color)}
\label{fig:result/gen_imgs_residual}
\end{figure}

\noindent \textbf{Can the model generate realistic images?} The trained model generates realistic looking medical images (second row in Figure~\ref{fig:result/gen_imgs_residual}) that are conditioned by sampling from latent representations $\mathbf{z}$, which are found through our mapping approach, described in Section~\ref{sec:methods:mapping}. In the case of normal image patches (see first and second block in Figure~\ref{fig:result/gen_imgs_residual}), our model is able to generate images that are visually similar to the query images (first row in Figure~\ref{fig:result/gen_imgs_residual}). But in the case of anomalous images, the pairs of input images and generated images show obvious intensity or textural differences (see third block in Figure~\ref{fig:result/gen_imgs_residual}). The t-SNE embedding (Figure \ref{fig:method/gan}(b)) of normal and anomalous images in the feature representation of the last convolution layer of the discriminator that is utilized in the \textit{discrimination loss}, illustrates the usability of the discriminator's features for anomaly detection and suggests that our AnoGAN learns a meaningful manifold of normal anatomical variability.

\begin{figure}[tp]
  \centering
     \includegraphics[width=1\textwidth]{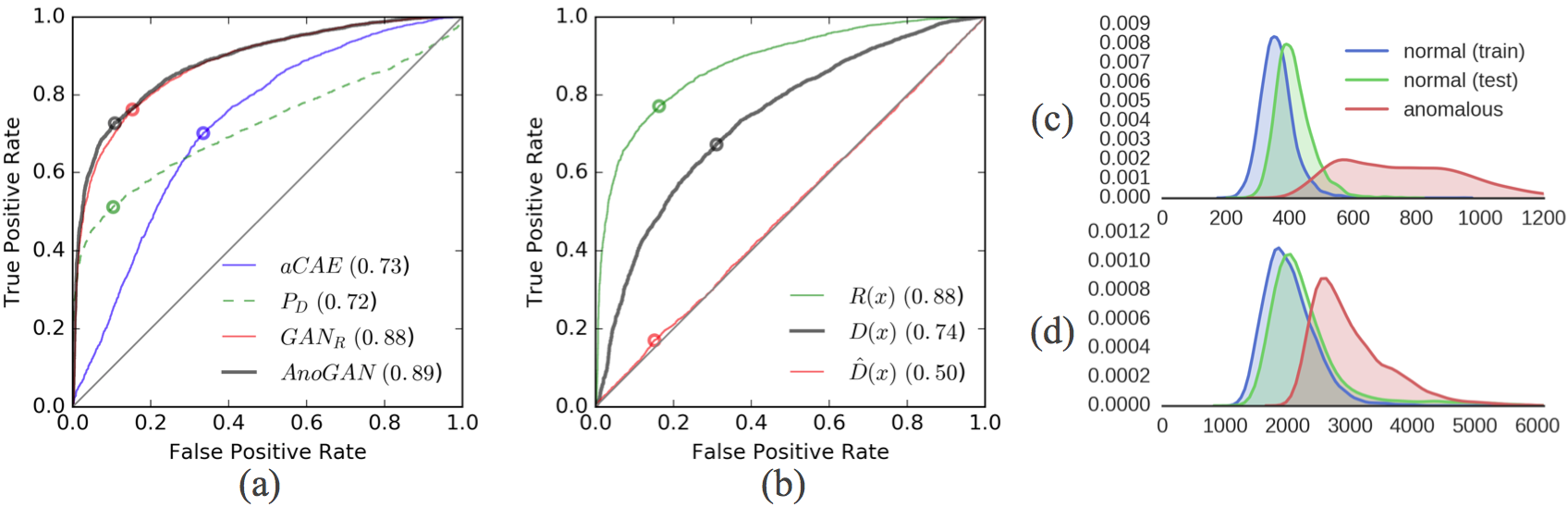}
  \caption{Image level anomaly detection performance and suitability evaluation. (a) Model comparison: ROC curves based on \textit{aCAE} (blue), $GAN_R$ (red), the proposed \textit{AnoGAN} (black), or on the output $P_D$ of the trained discriminator (green).
(b) Anomaly score components: ROC curves based on the \textit{residual score} $R(\mathbf{x})$ (green), the \textit{discrimination score} $D(\mathbf{x})$ (black), or the \textit{reference discrimination score} $\hat{D}(\mathbf{x})$ (red).
(c) Distribution of the \textit{residual score} and (d) of the \textit{discrimination score}, evaluated on normal images of the training set (blue) or test set (green), and on images extracted from diseased cases (red).}
\label{fig:result/roc}

\vspace{-5mm}
\end{figure}

\noindent \textbf{Can the model detect anomalies?} Figure~\ref{fig:result/roc}(b) shows the ROC curves for image level anomaly detection based on the 
\textit{anomaly score} $A(\mathbf{x})$, or on one of both components, on the \textit{residual score} $R(\mathbf{x})$, or on the \textit{discrimination score} $D(\mathbf{x})$. The corresponding area under the ROC curve (AUC) is specified in parentheses. In addition, the distributions of the \textit{residual score} $R(\mathbf{x})$ (Figure~\ref{fig:result/roc}(c)) and of the \textit{discrimination score} $D(\mathbf{x})$ (Figure~\ref{fig:result/roc}(d)) over normal images from the training set and test set or over images extracted from diseased cases show that both components of the proposed \textit{adversarial score} are suitable for the classification of normal and anomalous samples. Figure~\ref{fig:result/gen_imgs_residual} shows pixel-level identification of anomalies in conjunction with pixel-level annotations of retinal fluid, which demonstrate high accuracy. Last column in Figure~\ref{fig:result/gen_imgs_residual} demonstrates that the model successfully identifies additional retinal lesions, which in this case correspond to hyperreflective foci (HRF). On image level, the red and yellow bars in Figure~\ref{fig:result/gen_imgs_residual} demonstrate that our model successfully identifies every example image from diseased cases of the test set as beeing anomalous based on the \textit{residual score} and the \textit{discrimination score}, respectively.

\noindent \textbf{How does the model compare to other approaches?}
We evaluated the anomaly detection performance of the $GAN_R$, the \textit{aCAE} and the \textit{AnoGAN} on image-level labels. The results are summarized in Table~\ref{tab:result/clinic_stats_compare} and the corresponding ROC curves are shown in Figure~\ref{fig:result/roc}(a).
Although \textit{aCAEs} simultaneously yield a generative model and a direct mapping to the latent space, which is advantageous in terms of runtimes during testing, this model showed worse performance on the anomaly detection task compared to the \textit{AnoGAN}. It turned out that \textit{aCAEs} tend to over-adapt on anomalous images.
Figure~\ref{fig:result/roc}(b) demonstrates that anomaly detection based on our proposed \textit{discrimination score} $D(\mathbf{x})$ outperforms the \textit{reference discrimination score} $\hat{D}(\mathbf{x})$. Because the scores for the detection of anomalies are directly related to the losses for the mapping to latent space, these results give evidence that our proposed \textit{discrimination loss} $\mathcal{L}_{D}(\mathbf{z})$ is advantageous compared to the discrimination loss $\mathcal{L}_{\hat{D}}(\mathbf{z})$.
Nevertheless, according to the AUC, computed based on the \textit{anomaly score}, the \textit{AnoGAN} and the $GAN_R$ show comparable results (Figure~\ref{fig:result/roc}(a)). This has to be attributed to the good performance of the \textit{residual score} $R(\mathbf{x})$. A good anomaly detection performance (cf. $P_D$ in Figure~\ref{fig:result/roc}(a) and Table~\ref{tab:result/clinic_stats_compare}) can be obtained when the mapping to the latent space is skipped and a binary decision is derived from the discriminator output, conditioned directly on the query image.

\begin{table}[tp]
\centering
\caption{Clinical performance statistics calculated at the Youden's index of the ROC curve and the corresponding AUC based on the \textit{adversarial score} $A(\mathbf{x})$ of our model (\textit{AnoGAN}) and of the \textit{aCAE}, based on the \textit{reference adversarial score} $\hat{A}(\mathbf{x})$ utilized by $GAN_R$, or based directly on the output of the DCGAN ($P_D$).}
    \begin{tabular}
    { | @{}L{.2\textwidth} | C{0.15\textwidth} | C{0.15\textwidth} | C{0.15\textwidth} | C{0.15\textwidth} | C{0.15\textwidth} @{} | }
     \hline
 ~ & Precision & Recall & Sensitivity & Specificity & AUC \\ \hline
 \, aCAE    &0.7005 	&0.7009	&0.7011	&0.6659  &0.73\\ \hline
 \, $P_D$    &0.8471 &0.5119	&0.5124	&0.8970  &0.72\\ \hline
 \, $GAN_R$  &0.8482	&0.7631	&0.7634   &0.8477 &0.88\\ \hline
  \, AnoGAN &0.8834	&0.7277	&0.7279   &0.8928 &0.89\\
  \hline
    \end{tabular}
\normalsize
\label{tab:result/clinic_stats_compare}

\vspace{-5mm}
\end{table}

%% file: discussion.tex
\section{Conclusion} 
We propose anomaly detection based on deep generative adversarial networks. By concurrently training a generative model and a discriminator, we enable the identification of anomalies on unseen data based on unsupervised training of a model on healthy data. Results show that our approach is able to detect different known anomalies, such as retinal fluid and HRF, which have never been seen during training. Therefore, the model is expected to be capable to discover novel anomalies. While quantitative evaluation based on a subset of anomaly classes is limited, since false positives do not take novel anomalies into account, results demonstrate good sensitivity and the capability to segment anomalies. Discovering anomalies at scale enables the mining of data for marker candidates subject to future verification. In contrast to prior work, we show that the utilization of the residual loss alone yields good results for the mapping from image to latent space, and a slight improvement of the results can be achieved with the proposed adaptations.

%% file: 79.bbl
\begin{thebibliography}{10}

\bibitem{del2016discriminative}
Del~Giorno, A., Bagnell, J.A., Hebert, M.:
\newblock A discriminative framework for anomaly detection in large videos.
\newblock In: ECCV, Springer (2016)  334--349

\bibitem{matteoli2014overview}
Matteoli, S., Diani, M., Theiler, J.:
\newblock An overview of background modeling for detection of targets and
  anomalies in hyperspectral remotely sensed imagery.
\newblock IEEE Journal of Selected Topics in Applied Earth Observations and
  Remote Sensing \textbf{7}(6) (2014)  2317--2336

\bibitem{carrera2015detecting}
Carrera, D., Boracchi, G., Foi, A., Wohlberg, B.:
\newblock Detecting anomalous structures by convolutional sparse models.
\newblock In: 2015 International Joint Conference on Neural Networks (IJCNN),
  IEEE (2015)  1--8

\bibitem{erfani2016high}
Erfani, S.M., Rajasegarar, S., Karunasekera, S., Leckie, C.:
\newblock {High-dimensional and large-scale anomaly detection using a linear
  one-class SVM with deep learning}.
\newblock Pattern Recognition \textbf{58} (2016)  121--134

\bibitem{pimentel2014review}
Pimentel, M.A., Clifton, D.A., Clifton, L., Tarassenko, L.:
\newblock A review of novelty detection.
\newblock Signal Processing \textbf{99} (2014)  215--249

\bibitem{venhuizen2015automated}
Venhuizen, F.G., van Ginneken, B., Bloemen, B., van Grinsven, M.J., Philipsen,
  R., Hoyng, C., Theelen, T., S{\'a}nchez, C.I.:
\newblock Automated age-related macular degeneration classification in oct
  using unsupervised feature learning.
\newblock In: SPIE Medical Imaging, International Society for Optics and
  Photonics (2015)  94141I--94141I

\bibitem{schlegl2015ipmi}
Schlegl, T., Waldstein, S.M., Vogl, W.D., Schmidt-Erfurth, U., Langs, G.:
\newblock Predicting semantic descriptions from medical images with
  convolutional neural networks.
\newblock In: {International Conference on Information Processing in Medical
  Imaging}. Volume~24., Springer (2015)  437--448

\bibitem{seebock2016identifying}
Seeb{\"o}ck, P., Waldstein, S., Klimscha, S., Gerendas, B.S., Donner, R.,
  Schlegl, T., Schmidt-Erfurth, U., Langs, G.:
\newblock Identifying and categorizing anomalies in retinal imaging data.
\newblock NIPS 2016 MLHC workshop. preprint arXiv:1612.00686 (2016)

\bibitem{goodfellow2014generative}
Goodfellow, I., Pouget-Abadie, J., Mirza, M., Xu, B., Warde-Farley, D., Ozair,
  S., Courville, A., Bengio, Y.:
\newblock Generative adversarial nets.
\newblock In: Advances in Neural Information Processing Systems. (2014)
  2672--2680

\bibitem{denton2015deep}
Denton, E.L., Chintala, S., Fergus, R.,  et~al.:
\newblock Deep generative image models using a laplacian pyramid of adversarial
  networks.
\newblock In: Advances in neural information processing systems. (2015)
  1486--1494

\bibitem{donahue2016adversarial}
Donahue, J., Kr{\"a}henb{\"u}hl, P., Darrell, T.:
\newblock Adversarial feature learning.
\newblock arXiv:1605.09782 (2016)

\bibitem{radford2015unsupervised}
Radford, A., Metz, L., Chintala, S.:
\newblock Unsupervised representation learning with deep convolutional
  generative adversarial networks.
\newblock arXiv:1511.06434 (2015)

\bibitem{yeh2016semantic}
Yeh, R., Chen, C., Lim, T.Y., Hasegawa-Johnson, M., Do, M.N.:
\newblock Semantic image inpainting with perceptual and contextual losses.
\newblock arXiv:1607.07539 (2016)

\bibitem{salimans2016improved}
Salimans, T., Goodfellow, I., Zaremba, W., Cheung, V., Radford, A., Chen, X.:
\newblock {Improved techniques for training GANs}.
\newblock In: Advances in Neural Information Processing Systems. (2016)
  2226--2234

\bibitem{garvin2009automated}
Garvin, M.K., Abr{\`a}moff, M.D., Wu, X., Russell, S.R., Burns, T.L., Sonka,
  M.:
\newblock Automated 3-{D} intraretinal layer segmentation of macular
  spectral-domain optical coherence tomography images.
\newblock Transactions on Medical Imaging, IEEE \textbf{28}(9) (2009)
  1436--1447

\bibitem{DBLP:journals/corr/PathakKDDE16}
Pathak, D., Kr{\"{a}}henb{\"{u}}hl, P., Donahue, J., Darrell, T., Efros, A.A.:
\newblock Context encoders: Feature learning by inpainting.
\newblock CoRR \textbf{abs/1604.07379} (2016)

\bibitem{kingma2014adam}
Kingma, D., Ba, J.:
\newblock Adam: A method for stochastic optimization.
\newblock arXiv:1412.6980 (2014)

\bibitem{tensorflow2015-whitepaper}
Abadi, M., Agarwal, A., Barham, P., Brevdo, E., Chen, Z., Citro, C., Corrado,
  G.S., Davis, A., Dean, J., Devin, M., Ghemawat, S., Goodfellow, I., Harp, A.,
  Irving, G., Isard, M., Jia, Y., Jozefowicz, R., Kaiser, L., Kudlur, M.,
  Levenberg, J., Man\'{e}, D., Monga, R., Moore, S., Murray, D., Olah, C.,
  Schuster, M., Shlens, J., Steiner, B., Sutskever, I., Talwar, K., Tucker, P.,
  Vanhoucke, V., Vasudevan, V., Vi\'{e}gas, F., Vinyals, O., Warden, P.,
  Wattenberg, M., Wicke, M., Yu, Y., Zheng, X.:
\newblock {TensorFlow}: Large-scale machine learning on heterogeneous systems
  (2015) Software available from tensorflow.org.

\end{thebibliography}
